\documentclass{article}

\PassOptionsToPackage{numbers}{natbib}

 \usepackage[sglblindworkshop, final]{neurips_2026}
\usepackage{graphicx}
\usepackage{amsmath}

\workshoptitle{XAI4Science: Knowledge Discovery and Trust through Interpretable Foundation Models}

\usepackage[utf8]{inputenc} %
\usepackage[T1]{fontenc}    %
\usepackage{hyperref}       %
\usepackage{url}            %
\usepackage{booktabs}       %
\usepackage{amsfonts}       %
\usepackage{nicefrac}       %
\usepackage{microtype}      %
\usepackage{xcolor}         %
\usepackage[nameinlink,capitalize]{cleveref}
\crefname{appendix}{App.}{Appendices}
\Crefname{appendix}{App.}{Appendices}

\usepackage{float}
\usepackage{caption}
\usepackage{enumitem}
\usepackage{tikz}

\usetikzlibrary{arrows.meta,positioning,fit,backgrounds,calc}

\newcommand{\N}{\mathcal{N}}
\newcommand{\R}{\mathbb{R}}
\newcommand{\Rval}{R^{\mathrm{val}}}
\newcommand{\Ratt}{R^{\mathrm{att}}}

\usepackage{soul}

\title{Value or Attention: Investigating Layer-wise Relevance Propagation for Graph Attention Networks}

\author{%
  Ken Schreiber\textsuperscript{1} \quad
  René P.~Klausen\textsuperscript{1} \quad
  Sebastian Lapuschkin\textsuperscript{1,2} \\
  \textbf{Wojciech Samek\textsuperscript{1,3,4} \quad
  Jonas Naujoks\textsuperscript{1,\dag}} \\[6pt]
  \textsuperscript{1}Department of Artificial Intelligence, Fraunhofer Heinrich Hertz Institute, Berlin, Germany \\
  \textsuperscript{2}Centre of eXplainable Artificial Intelligence, Technological University Dublin, Ireland \\
  \textsuperscript{3}Department of Electrical Engineering, Technische Universität Berlin, Germany \\
  \textsuperscript{4}BIFOLD -- Berlin Institute for the Foundations of Learning and Data, Berlin, Germany \\[4pt]
  \textsuperscript{\dag}Corresponding author: \texttt{jonas.naujoks@hhi.fraunhofer.de}
}

\begin{document}

\maketitle

\begin{abstract}
Graph Attention Networks (GATs) are increasingly used in domains where scientists need to read a model's reasoning, from medicine to materials science, yet the routing they learn remains opaque to most explainers. Layer-wise Relevance Propagation (LRP) follows the model's computation, but in an attention layer the relevance reaching a node is the sum of two mechanically different quantities: its content relevance through the value path and its routing relevance through the attention path. Existing rules either drop one or add them into a single number. We write a single product rule with free weights on the two factors that recovers current approaches as special cases. The value path measures what content a node supplied against a zero baseline; the attention path measures how the node shaped what was passed relative to its neighborhood, and its sign records whether the node's features moved the gating the way the prediction demanded. On a ground-truth classification task, we show that the same node receives negative, positive and near-zero relevance under three settings of the rule. The separated relevance maps yield a hypothesis about the model: it gains by not reading that node. We give reading rules for both paths and argue that they should be interpreted separately.
\end{abstract}

\section{Introduction}

From particles to planets, social networks to logical formulas, many scientific relations can naturally be described as graphs. Graph Neural Networks (GNNs) \citep{scarselli2009GNN} learn on such graphs directly by passing messages between neighboring nodes. How a node aggregates its neighbors' messages is their central design choice: Graph Convolutional Networks (GCNs) \citep{kipf2017GCN} use coefficients fixed by the graph structure, while Graph Attention Networks (GATs) allow input-dependent routing \citep{brody2022gat2,velickovic2018gat}. GATs are deployed in high-stakes domains such as biomedical systems \citep{baranwal2022struct2graph,hu2024gatbio,otal2025gatgene}, electronic health records \citep{osboll2024gathealth,piya2024gathealth}, drug discovery \citep{lv2024gatdrug}, and materials science \citep{schmidt2021gatmaterials}, where inspectability of a model's reasoning is needed for trust and scientific insight. In these settings, an explanation proposes a hypothesis about which nodes mattered and why, which requires that the importance assigned to a node have a single, stateable meaning.

For GNNs, most explainability approaches are perturbation- or surrogate-based \citep{huang2023GraphLime, luo2020PGExplainer,vu2020PGMExplainer,ying2019GNNExplainer,yuan2021subgraphx}: they score a node by the effect of intervening on it. In an attention layer, however, a node influences the prediction through two mechanisms: the message it sends and the attention weights it helps compute. We refer to these as the \emph{value path} and the \emph{attention path}. Any intervention at the input moves both, and black-box methods thus cannot distinguish the two mechanisms. Layer-wise Relevance Propagation (LRP) \citep{bach2015pixelwiseLRP}, and its graph extension GNN-LRP \citep{schnake2022relevantwalks}, instead decompose the prediction along the model's computation and can attribute to each mechanism through the model's internals. There is no canonical way to split the relevance across the two paths. The question has been studied in the context of Transformers: CP-LRP \citep{ali2022CPLRP} keeps only the value path, while AttnLRP \citep{achtibat2024AttnLRP} distributes evenly across both paths. These rules trade places across metrics and model families \citep{arras2025LLM,you2025blackboxNLP}. Neither rule considers the two paths separately, which only matters if they carry different information.

We show that they do. By writing the relevance through one GATv2 layer as a product rule with free weights on the two factors, we find that the two paths answer different questions: the value path says how much each node's content contributed; the attention path, how each node shaped which content was routed. Existing rules are special cases of this formulation. Dropping either path omits information, while collapsing them into a single attribution entangles distinct stories. On a graph-classification task with known ground truth \textsc{BA-House} \citep{luo2020PGExplainer}, the same node receives a positive, negative or near-zero relevance depending on the setting of the rule. We therefore recommend that the two paths be interpreted individually. We show how this can lead to deeper mechanistic understanding, which we argue increases trust and opens up potential for scientific discovery.

\section{Background and related work}

\paragraph{Graph Attention Networks}

We study GATv2 \citep{brody2022gat2}, the dynamic-attention variant of GAT. A GATv2 layer updates node $i$ from its neighborhood $\N(i)$ by message passing 
\begin{equation}
  y_i=\!\!\sum_{j\in\N(i)}\!\!\alpha_{ij}m_j,
\label{eq:gatv2mp}
\end{equation}
where messages from all source nodes $j$ are aggregated over a single destination $i$. The attention parameter $\alpha_{ij}$ and message $m_j$ are calculated as
\begin{equation}
  m_j = W_l x_j,\qquad
  s_{ij} = a^\top\sigma\!\left(W_l x_j + W_r x_i + W_e e_{ij}\right), \qquad
  \alpha_{ij} = \mathrm{softmax}_j(s_{ij}).
\label{eq:gatv2}
\end{equation}
where $\sigma$ is the LeakyReLU, $e_{ij}$ optional edge attributes, $W_l,W_r,W_e$ learned source-, destination- and edge-projections, $a$ a learned score vector, and $x_j, x_i$ source- and destination-node features from the layer input $x$. A message passing step differs from a Transformer attention layer in two ways: (i) the embeddings: $W_lx_j$ is \emph{both} the message and the source-side input to $s_{ij}$, and optional edge attributes $e_{ij}$ enter \emph{only} the attention path; (ii) the logit $s_{ij}$ calculation, which mimics a feed-forward layer compared to the Transformer's normalized matrix multiplication $s_{ij}=q_i^\top k_j/\sqrt{d_k}$. 

\paragraph{Layer-wise Relevance Propagation (LRP)}
LRP \citep{bach2015pixelwiseLRP} is a well-established and efficient explanation method settled between gradient- and decomposition-based attribution: originally motivated as a rule-based, layer-by-layer redistribution under a conservation principle, modern rule formulations often follow a Deep Taylor Decomposition motive \citep{montavon2017DTD} and can be implemented as a \emph{modified} Gradient$\times$Input (G$\times$I) pass \citep{ali2022CPLRP,arras2025LLM}. Because LRP rules rest on local linearizations, a central objective of LRP research is deriving faithful rules for the non-linear operations that violate this assumption. 

\paragraph{Related work} 
As a special extension for GNNs, a higher-order method was developed that nests first-order explanations into relevant walks \citep{schnake2022relevantwalks}. However, they build on the first-order rules, where GAT message passing faces the same non-linearities that are unsettled for Transformer attention: the bilinear product and the softmax function. 
\citet{ali2022CPLRP} motivate CP-LRP by showing that G$\times$I breaks conservation at the softmax whenever its logits and the inputs covary; treating the attention weights as constants removes that term. \citet{achtibat2024AttnLRP} maintain that discarding how the routing was produced costs faithfulness. Their rule, motivated by Deep Taylor Decomposition and Shapley decomposition at a zero baseline, preserves conservation at the bilinear product via uniform relevance split and accepts the softmax as a legitimate relevance sink. In the recent literature, both rules achieve mixed results across model families and metrics \citep{arras2025LLM,you2025blackboxNLP} --- the cause is stated as an open question \citep{arras2025LLM}.

\section{Methodology} \label{sec:theory}

We suggest that these results reflect a genuine difference in what the two rules measure, not necessarily a shortcoming of either. We argue that deep understanding of how both relevance paths function individually will lead to mechanistic understanding beyond what's possible with a single reading. This consequently gives insights into the different questions that CP-LRP and AttnLRP answer. We make this separate reading precise for GAT message passing.

\subsection{The two paths gradient rule}

For a single message passing step (\cref{eq:gatv2mp,eq:gatv2}), we write the \emph{modified} gradient of an output feature $y_i$ with regard to the input vector $x$ with an adapted product rule with a set of free weights $\mathbf{w} = (w_{\mathrm{val}}, w_{\mathrm{att}})$, so that the existing methods appear as special cases:
\begin{equation}
  \left.\frac{\partial y_i}{\partial x}\right|_{\mathbf{w}}
  =\sum_{j\in\N(i)}\Big[\;
     \underbrace{w_{\mathrm{val}}\,\alpha_{ij}\,\tfrac{\partial m_j}{\partial x}}_{\text{value path}}
     \;+\;
     \underbrace{w_{\mathrm{att}}\,m_j\,\tfrac{\partial \alpha_{ij}}{\partial x}}_{\text{attention path}}
  \;\Big].
  \label{eq:split}
\end{equation}
We recover the formulations for G$\times$I at $\mathbf{w}=(1,1)$, the CP-LRP rule at $\mathbf{w}=(1,0)$, the AttnLRP rule at $\mathbf{w}=(\frac{1}{2},\frac{1}{2})$, and inspect the attention path individually at $\mathbf{w}=(0,1)$. In deeper networks the choice recurs at each layer and rules face the question of the downstream implications of relevance aggregation between the paths. A caveat to the distinctiveness of the two paths is that in GATv2 the two paths share a derivative, which is further discussed in \cref{app:gatatt}.

\subsection{Two types of object}\label{ssec:types}

Let $g_i$ be the incoming gradient signal at destination node $i$ for the explained scalar output. We recover the value and attention path relevances by multiplying with the activation. For the full edge-level algebra, we refer to \cref{app:derivation}.

\paragraph{The value path as a content share}
At $\mathbf{w}=(1,0)$ only the value term of \cref{eq:split} remains, and the edge relevance is $r^{\mathrm{val}}_{ij}=\alpha_{ij}\,g_i^\top m_j$. Summing over the destinations that $j$ sends to gives
$R^{\mathrm{val}}_{j}=\sum_{i\,:\,j\in\N(i)} r^{\mathrm{val}}_{ij}$.
With fixed routing, this is a linear function with factors $\alpha_{ij}\ge 0$. Layer-locally, at fixed $g_i$, the ablation effect is exact: zeroing node $j$'s message removes exactly $r^{\mathrm{val}}_{ij}$ from the relevance $g_i^\top y_i$ that destination $i$ received, and exactly $R^{\mathrm{val}}_{j}$ across all of $j$'s outgoing edges (\cref{app:exact}).

\paragraph{The attention path as a routing influence}
At $\mathbf{w}=(0,1)$ the attention path is isolated. Differentiating through the softmax leaves the relevance on the logit $s_{ij}$,
\begin{equation}
  r^{\mathrm{att}}_{ij}
  = r^{\mathrm{val}}_{ij}-\alpha_{ij}\!\!\sum_{k\in\N(i)}\!\! r^{\mathrm{val}}_{ik},
  \qquad
  \sum_{j\in\N(i)} r^{\mathrm{att}}_{ij}=0 ,
  \label{eq:ratt}
\end{equation}
which is the value path relevance but subtracts from each edge its share of the neighborhood total: it measures the effect of re-routing relevance among neighbors. Withholding attention from one neighbor necessarily grants it to the others, and relevance can only exist in relation (\cref{app:derivation}).

Reaching an input node takes one further step. The logit is computed from the source's features $x_j$ and the destination's features $x_i$. A node $n$ therefore collects relevance in two roles,
\begin{equation}
    R^{\mathrm{att}}_{n}
    =\underbrace{\sum_{i\,:\,n\in\N(i)} r^{\mathrm{att}}_{in}\;u_{in}}_{n \text{ as source}}
    \;+\;
    \underbrace{\sum_{j\in\N(n)} r^{\mathrm{att}}_{nj}\;v_{nj}}_{n \text{ as destination}},
    \qquad
    \begin{aligned}
        u_{in} &= x_n^\top W_l^\top p_{in},\\
        v_{nj} &= x_n^\top W_r^\top p_{nj},
    \end{aligned}
  \label{eq:rattnode}
\end{equation}
with $p_{ij} = \sigma'(W_l x_j + W_r x_i + W_e e_{ij}) \odot a$ the local linearization of the scorer (\cref{app:node}). Both factors in each product are signed and meet at the logit from opposite directions: $r^{\mathrm{att}}_{ij}$ is a \emph{demand} from above --- which way the target wants this edge scored --- while $u_{in}$ and $v_{nj}$ are \emph{supplies} from below, how much of that logit $n$'s own features produced. The product is positive when demand and supply agree: the edge relevance $r^{\mathrm{att}}_{ij}$ says score this edge higher and $n$'s features push the logit up, or it says score it lower and they push it down.

The roles differ in what survives aggregation. The destination role sum ranges over $\N(n)$, the same set on which the attention parameters are normed, but it does not inherit the edge relevances' zero-sum (\cref{eq:ratt}) property due to the non-linear LeakyReLU. It asks how $n$'s features affected the way it distributed attention to its neighbors within a constrained attention fund; its magnitude is set by how much those features separate one neighbor from another. The source role sum ranges over multiple constrained destination neighborhoods and is itself unconstrained. It measures how $n$'s features affected the way it was read by its neighbors; its magnitude is set by how much the neighborhoods that read $n$ differ in what they demand of it.

\paragraph{Consequences for interpreting attributions}
Value path and attention path attributions read differently. A value path content share is measured against a $0$ baseline, it vanishes when a node contributes nothing. The sign means: $j$'s content moved the target this way. Positive attributions have a clear reading: \emph{my content helped the target by this much}, where \emph{helping} means pushing the class logit up. CP-LRP will assign this relevance, but miss all information from the attention calculation. An attention path routing influence has no fixed reference, its magnitude depends on contrast within the neighborhood, it vanishes when all content matters the same, regardless of what they contribute. Its sign measures a demand-supply agreement: in the source role a positive sign means \emph{skip me, and skipping me helps} or \emph{read me, and reading me helps}. For a flowchart of all interpretations find \cref{app:flows}. AttnLRP will combine the information of both paths, but \emph{content helped} and \emph{steering helped} become a single number with overlapping magnitude meanings and sign readings.

\section{Experiments} \label{sec:experiments}

\begin{figure}[tb]
    \centering
    \includegraphics[width=0.8\textwidth]{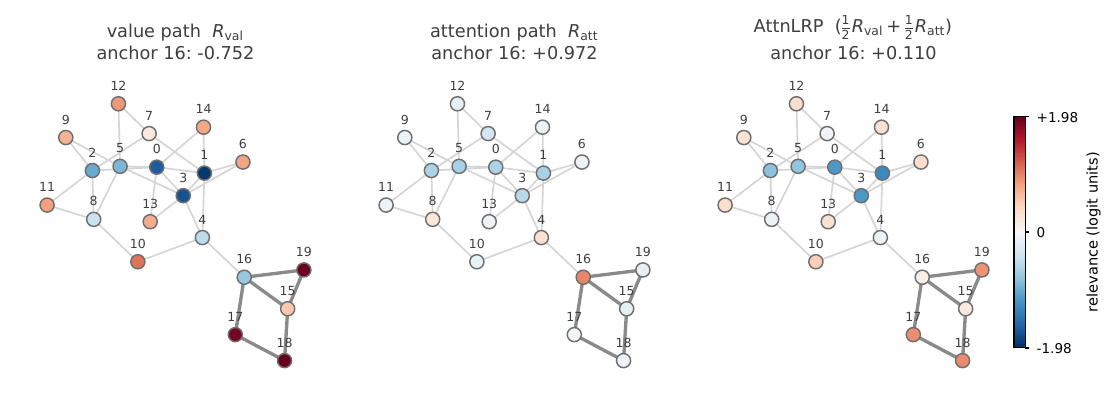}
    \caption{Node relevance on one \textsc{BA-House} \citep{luo2020PGExplainer} graph, red positive, blue negative.}
    \label{fig:bahex}
\end{figure}

\cref{sec:theory} argues algebraically that the two paths answer different questions. In order to make these readings tangible, we present a case study as a proof of concept. It centers on a demonstrative instance where each path tells a coherent, yet distinct story about the same prediction on a singular node. For further experiments on a toy dataset and a real-world motivation, we refer to \cref{app:blind}. For the case study, we evaluate the value path (CP-LRP), the attention path, and AttnLRP ($\mathbf{w} \in \{(1,0)$, $(0,1)$, $(\frac{1}{2},\frac{1}{2})\}$ in \cref{eq:split}) at a single GATv2 layer on \textsc{BA-House}, a graph-classification special case of the synthetic BA-2Motifs benchmark \citep{luo2020PGExplainer}, chosen for its interpretable ground truth. A positive graph contains a house motif consisting of two \emph{floor} nodes, two \emph{eave} nodes, and one \emph{rooftop} node (\cref{fig:bahex}). The attachment node becomes the \emph{anchor}: a floor or rooftop anchor reaches degree three, an eave anchor reaches degree four. The node embeddings are their one-hot encoded degree.

\cref{fig:bahex} shows the three attribution maps on one graph with a degree-four anchor node $16$ and highlights disagreement on the same node: the value path assigns negative and the attention path positive relevance, the sum returns it to near zero. On the base-graph hubs ($0,1,2,3,5$) the two paths agree in sign. The trend between the two paths attributions can be similar or opposite, depending on the nodes and neighborhoods, and a combined reading entangles this information.

\begin{figure}[bt]
    \centering
    \includegraphics[width=0.8\textwidth]{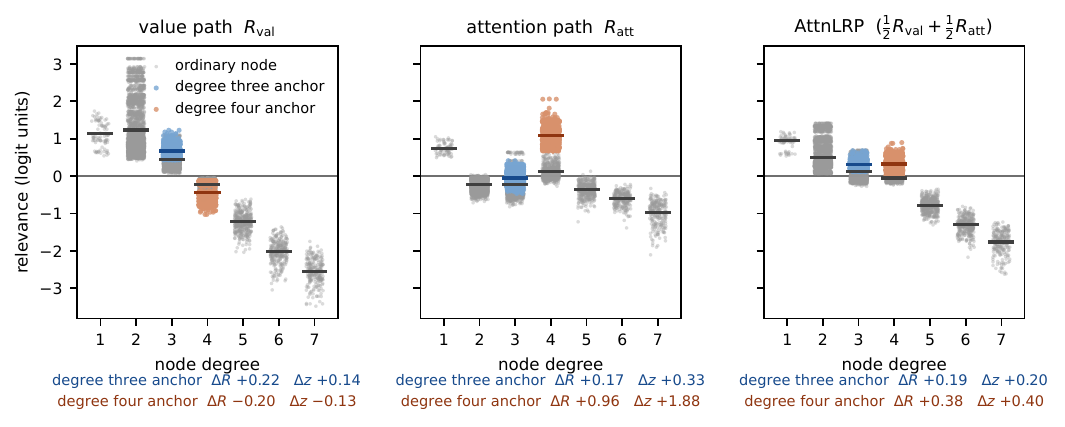}
    \caption{Node relevance for the house logit against degree, all nodes of $600$ positive graphs across 5 initializations. The 4.6\% of nodes with degree higher than 7 are omitted here for clarity; \cref{fig:bahdegreefull} shows the full range. Grey: ordinary (non-anchor) nodes, from both the BA graph and the house; black bars: their per-degree mean. Blue / orange: anchor nodes of degree three / four; dark blue / dark orange bars: their means. $\Delta R$: mean anchor relevance minus mean ordinary-node relevance at the same degree. $\Delta z$: $\Delta R$ divided by the map's standard deviation over all nodes, including those not drawn, which makes it independent of each method's overall scale (per initialization, then averaged).}
    \label{fig:bahdegree}
\end{figure}

\cref{fig:bahdegree} shows that this is not an isolated case. 100\% of evaluated degree-four nodes receive negative value path relevance (max $-0.031$), which, by \cref{ssec:types}, reads as a negative contribution to the class logit. Read as content alone, the negative attribution is reasonable, because house nodes are naturally degree two or degree three, the only case a degree-four node can belong to a house is as an anchor node at the eave: only $18.2\%$ of degree-four nodes belong to a house ($41.6\%$ for degree-three and $27.9\%$ for degree-two). 
The attention path attributes them positive relevance and lifts degree-four anchors' relevance substantially ($\Delta z = +1.88$) relative to their main graph counterparts. This signal sits in the source role: across the evaluation set the source term carries $93.4\%$ of $|R^{\mathrm{att}}|$ at degree-four anchor nodes. Following \cref{ssec:types} routing influence logic, the positive relevance means the \emph{skip me, and skipping me helps} case: 
For node 16 (\cref{fig:bahex}), 100\% of its 5 sender edges have both \emph{demand} $r^{\mathrm{att}}<0$ and \emph{supply} $u_{in}<0$ (\cref{eq:rattnode}). 

Combining both maps flattens the separation to $\Delta z=+0.40$ (\cref{fig:bahdegree}, right) and node $16$ to near zero (\cref{fig:bahex}, right). This is not ``wrong'': in this example, one could even argue that the combined path explanation finds the best house representation.
What is lost is the disentanglement between interpretations. \emph{This content helped} and \emph{this steering helped} (\cref{ssec:types}) get reported as a single value, and a small entry no longer distinguishes a node that neither contributes value nor routing from a node whose two roles point in opposite directions with high influence. 

In this case study, the value path reports ``degree 4 nodes contribute negatively to the classification logit'', the attention path reports ``it is important that degree 4 nodes don't send their information, but especially in high value neighborhoods'', and their combination says ``in some way, degree 4 nodes are important to the prediction''. It shows that close, differentiated inspection between the content contribution and routing roles can lead to meaningful insights about how the model comes to conclusions. This should also be kept in mind when evaluating models with metrics. For example, some graph explainability metrics mask messages that get sent to and from a node; they can't evaluate nodes that contribute by not being read in the first place (\cref{app:expfid}).

\section{Conclusion}

Graph Attention Networks (GATs) are applied in domains where explanations must support trust and scientific insights.
A standard way to explain GNNs is to intervene at the input, but such an intervention moves both a node's message and its attention weights. Separating the two requires attributions at the model's internals, which Layer-wise Relevance Propagation (LRP) provides for all nodes with one backward pass per path. We have shown that in an attention layer, the relevance reaching a node is the sum of a value path share and an attention path routing influence. These two differ in baseline, in conservation, and interpretation. Existing rules either drop one or add them into a single number.

Since these rules are settings of one product rule, considering the paths separately is straightforward and gives each a definite meaning. On \textsc{BA-House}, where the ground truth is known, our case study shows the node attaching the motif to the graph receives negative, positive and near-zero relevance under three different settings of the rule. We read them separately and show: the value path says the node's content contributes negatively to the predicted class and the attention path says the model's gating of it works in the model's favor.
Which rule to use therefore depends on what is asked of an explanation.
For a domain expert, the separated reading turns one ambiguous number into two statements that can be analyzed independently. In high-stakes domains in particular, the distinction matters for trust: a near-zero attribution because a node has little influence and a near-zero attribution because its counterproductive content is suppressed by the routing are different statements. Beyond trust, for a model that predicts well, the separate readings can yield more specific hypotheses about what it has learned. Tested against the data by domain experts, such hypotheses can contribute to scientific insight. 
We see this as a step towards a mechanistic understanding of both GATs by themselves and their attention mechanism, in particular, compared to attention-free GNNs.

\paragraph{Limitations and future work}

Our attention split is posed at a single attention layer, where it is exact. Furthermore, our main experiments use single-head models and synthetic data with known ground truth. In GATv2, the message and the source embedding share their projection, so information can be entangled even though the computation is decoupled. 

Next steps are to extend the experiments to real-world data, to tasks where attention is provably necessary, and to models with multiple attention heads and layers. The main open question is whether a combination exists that retains the information of the individual readings better than the sum used in AttnLRP, so that a more meaningful relevance propagates through preceding layers. Perturbation studies of node-level attention path attributions could guide this.

In principle, the different readings can be applied to other architectures that share the bilinear, softmax-normalized attention. To what extent the split accounts for the disagreement between CP-LRP and AttnLRP in Transformers remains open.

\section*{Acknowledgments}
This work was supported by the Fraunhofer Internal Programs (Fraunhofer) as grant ESPINN (PREPARE 40-08394).

\bibliography{references}
\bibliographystyle{plainnatnourl}

\appendix
\crefalias{section}{appendix}
\crefalias{subsection}{appendix}

\section{GATv2 vs. Transformer attention}
\label{app:gatatt}
A Transformer attention layer has separate projections for the input on the sender side: $W_k$ for the key that enters the attention computation, and $W_v$ for the value that gets sent.
\begin{equation}
y_i = \sum_{j} \operatorname{softmax}_j\!\left( \frac{q_i^\top k_j}{\sqrt{d_k}} \right) v_j,
\qquad q_i = W_q x_i,\quad k_j = W_k x_j,\quad v_j = W_v x_j .
\end{equation}
With $s$ the logit, the backward pass splits schematically into
\begin{equation}
\frac{\partial y}{\partial x} = \underbrace{\frac{\partial y}{\partial \alpha} \frac{\partial \alpha}{\partial s} \left( \frac{\partial s}{\partial q} \frac{\partial q}{\partial x} + \frac{\partial s}{\partial k} \frac{\partial k}{\partial x} \right)}_{\text{attention path}} + \underbrace{\frac{\partial y}{\partial v} \frac{\partial v}{\partial x}}_{\text{value path}}.
\end{equation}

In GATv2 message passing, the source node that sends the message has a single projection $W_l$ that enters the attention computation and gets transmitted as the message,
\begin{equation}
y_i = \sum_j \operatorname{softmax}_j \left( a^\top \operatorname{LeakyReLU}\left( W_r x_i + W_l x_j \right) \right) W_l x_j.
\end{equation}

\begin{equation}
\frac{\partial y}{\partial x} = \underbrace{\frac{\partial y}{\partial \alpha} \frac{\partial \alpha}{\partial s} \frac{\partial s}{\partial \hat z} \frac{\partial \hat z}{\partial z} \left( \frac{\partial z}{\partial q} \frac{\partial q}{\partial x} + \frac{\partial z}{\partial m} \frac{\partial m}{\partial x} \right)}_{\text{attention path}} + \underbrace{\frac{\partial y}{\partial m} \frac{\partial m}{\partial x}}_{\text{value path}} ,
\end{equation}
where, as in \cref{eq:gatv2} with edge attributes omitted for simplicity, $q_i = W_r x_i$ and $m_j = W_l x_j$ are the destination and source projections, $z_{ij} = q_i + m_j$ the pre-activation, $\hat z_{ij} = \sigma(z_{ij})$ its LeakyReLU, and $s_{ij} = a^\top \hat z_{ij}$ the logit.

In the GATv2 backward pass, both the attention path and the value path carry the factor $\partial m/\partial x = W_l$. This does not couple the decomposition. \cref{eq:split} is exact and its two terms are disjoint, so $w_{\mathrm{att}}=0$ removes the attention term entirely and $R^{\mathrm{val}}$ carries no routing contribution, and conversely. What the sharing does limit is how far \emph{agreement} between the two maps can be read as independent evidence, since the shared projection was fit under both objectives at once. It does not force that agreement: both readings express a node through the same vector $W_lx_n$, but score it against different covectors --- the upstream signal $g_i$ on the value path, the scorer's linearization $p_{in}$ in the source role (\cref{eq:rattnode}). In Transformer attention, the two paths share no such factor.

Empirically, the architecture control in \cref{app:bahctl} (\cref{fig:bahctl}), which compares GATv2 with \texttt{TransformerConv} (separate $W_k$, $W_v$) and an attention-free GCN, suggests that the shared projection has little effect, and that the architectural similarities of the bilinear product and softmax normalization outweigh the different embedding and logit computation, at least on our case study dataset. The shared projection does not change what the two maps report qualitatively.

\section{Edge-level algebra and the step to nodes}
\label{app:derivation}

\subsection{Edge relevances}
Let $T$ be the explained scalar and $g_i$ the signal that reaches the layer
output $y_i$: the standard gradient $\partial T/\partial y_i$ or its modified-gradient counterpart under any upstream rules. 
The value path edge relevance is one application of G$\times$I at the message, with
\begin{equation}
  r^{\mathrm{val}}_{ij}
  = m_j^\top \frac{\partial T}{\partial m_j}
  = m_j^\top \frac{\partial T}{\partial y_i}\frac{\partial y_i}{\partial m_j}
  = \alpha_{ij}\,g_i^\top m_j .
\end{equation}

For the attention path we need
\begin{align}
  \frac{\partial T}{\partial \alpha_{ij}} &= g_i^\top m_j,
  &
  \frac{\partial \alpha_{ik}}{\partial s_{ij}} &= \alpha_{ik}\left(\delta_{kj}-\alpha_{ij}\right),
\end{align}
so that
\begin{equation}
  \begin{aligned}
    r^{\mathrm{att}}_{ij}
    &= \frac{\partial T}{\partial s_{ij}}
     = \sum_{k\in\N(i)} \frac{\partial T}{\partial \alpha_{ik}}\frac{\partial \alpha_{ik}}{\partial s_{ij}}
     = \sum_{k\in\N(i)} g_i^\top m_k\,\alpha_{ik}\left(\delta_{kj}-\alpha_{ij}\right)\\
    &= \alpha_{ij}\,g_i^\top m_j - \alpha_{ij}\sum_{k\in\N(i)}\alpha_{ik}\,g_i^\top m_k
     = \alpha_{ij}\,g_i^\top\!\left(m_j-y_i\right)\\
    &= r^{\mathrm{val}}_{ij} - \alpha_{ij}\sum_{k\in\N(i)} r^{\mathrm{val}}_{ik}.
  \end{aligned}
  \label{eq:rattderiv}
\end{equation}
The middle form $\alpha_{ij}\,g_i^\top(m_j-y_i)$ can also be interpreted as a statement of what the map measures: the neighbor's content score against the attention-weighted level of its own neighborhood.

\subsection{Exactness of the value path ablation}
\label{app:exact}
With the attention weights frozen the layer is linear in the messages, so setting $m_j\to0$ gives $\Delta y_i=-\alpha_{ij}m_j$ with no remainder and no interaction with the other neighbors, whose coefficients do not depend on $m_j$. With $g_i$ fixed,
\begin{equation}
  \Delta\big(g_i^\top y_i\big) = -r^{\mathrm{val}}_{ij},
  \qquad
  \sum_{i\,:\,j\in\N(i)} \Delta\big(g_i^\top y_i\big) = -R^{\mathrm{val}}_j ,
\end{equation}
so G$\times$I at the edge level message is not an estimate of the ablation effect but the effect itself. The statement requires an intervention that leaves every $\alpha_{ij}$ untouched: masking the edge message qualifies, since the scorer never reads $m_j$, whereas zeroing $x_j$ or deleting the node moves the logits as well and perturbs both paths at once, as in an ordinary forward pass the routing is never frozen. 

The attention path cannot work this way: the softmax is nonlinear in the logits and its row normalization couples all neighbors of a destination.
 
\subsection{Row-sum cancellation}
\label{app:rowsum}
With $\sum_{j\in\N(i)}\alpha_{ij}=1$, the second term of \cref{eq:rattderiv} sums to the first,
\begin{align}
  \sum_{j\in\N(i)} r^{\mathrm{att}}_{ij}
  &= \sum_{j\in\N(i)} r^{\mathrm{val}}_{ij}
   - \Big(\sum_{j\in\N(i)}\alpha_{ij}\Big)\sum_{k\in\N(i)} r^{\mathrm{val}}_{ik}
   = 0 .
\end{align}

Attention is a fixed budget: what one neighbor is given is taken from the others. The question at the logit is not how much an edge delivers but whether it delivers more than the neighbors it competes with. A below-average neighbor is scored negative however much relevance its edge carries in total. Thus, ordering of attention path relevances across edges can be completely different from the value path, and it works on a different scale, depending on how different the neighborhood value path relevances are.

\subsection{From logits to nodes}
\label{app:node}

The relevance of \cref{eq:rattderiv} sits on $s_{ij}$. With $z_{ij} = W_l x_j + W_r x_i + W_e e_{ij}$ and $\sigma$ applied elementwise,
\begin{align}
  \frac{\partial s_{ij}}{\partial x_j} &= W_l^\top p_{ij},
  &
  \frac{\partial s_{ij}}{\partial x_i} &= W_r^\top p_{ij},
  &
  p_{ij} &= \sigma'(z_{ij}) \odot a .
\end{align}
Applying G$\times$I at $x_n$ and collecting the logits $x_n$ enters --- as a source in $s_{in}$ for each destination $i$, and as a destination in $s_{nj}$ for each of its own neighbors $j$ --- gives \cref{eq:rattnode}.

Two remarks. First, the source-side input to the scorer is $W_l x_j$, the same projection that produces the message; as the attention path explanation uses those weights in its gradient, it inherits information from the training of the value path. The projection still influences both paths separately, as different upstream relevance signals arrive at that projection. Second, with self-loops a node is its own neighbor and the self-edge contributes to both sums.

\section{Path reading flowchart}
\label{app:flows}
\cref{tab:signs} and \cref{tab:signtable} show the possible reading of the value and the attention path attributions.
\begin{table}[htbp]
\centering
\small
\setlength{\tabcolsep}{3pt}
\caption{Readings of a \textbf{positive} attribution. The value path does not branch; the
attention path branches by role and, within each role, by the direction on which demand
and supply agreed. The negative case is given in \cref{tab:signtable}.}
\label{tab:signs}
\begin{tabular}{@{}lccccc@{}}
\toprule
& value path & \multicolumn{4}{c}{attention path} \\
\cmidrule(lr){2-2}\cmidrule(l){3-6}
& & \multicolumn{2}{c}{$n$ as source} & \multicolumn{2}{c}{$n$ as destination} \\
\cmidrule(lr){3-4}\cmidrule(l){5-6}
node map & $R^{\mathrm{val}}_{n}$
         & \multicolumn{2}{c}{$\sum_{i\,:\,n\in\N(i)} r^{\mathrm{att}}_{in}u_{in}$}
         & \multicolumn{2}{c}{$\sum_{j\in\N(n)} r^{\mathrm{att}}_{nj}v_{nj}$} \\
edge term & $\alpha_{in}\,g_i^\top m_n$
          & \multicolumn{2}{c}{$r^{\mathrm{att}}_{in}\,u_{in}$}
          & \multicolumn{2}{c}{$r^{\mathrm{att}}_{nj}\,v_{nj}$} \\
reference & $0$ baseline
          & \multicolumn{2}{c}{neighborhoods reading $n$}
          & \multicolumn{2}{c}{$\N(n)$, contrasted twice} \\
scales with & content $m_n$
            & \multicolumn{2}{c}{contrast in demand}
            & \multicolumn{2}{c}{contrast in demand and in supply} \\
\midrule
supply (below) & ---
  & $u_{in}>0$ & $u_{in}<0$
  & $v_{nj}>0$ & $v_{nj}<0$ \\
 & & $n$ pushes logit up & $n$ pushes logit down
   & $n$ rates $j$ above & $n$ rates $j$ below \\
demand (above) & ---
  & $r^{\mathrm{att}}_{in}>0$ & $r^{\mathrm{att}}_{in}<0$
  & $r^{\mathrm{att}}_{nj}>0$ & $r^{\mathrm{att}}_{nj}<0$ \\
 & & score edge up & score edge down & shift toward $j$ & shift away from $j$ \\
\midrule
reading
& \emph{my content}  & \emph{read me,}    & \emph{skip me,}      & \emph{I favor this one,} & \emph{I pass this one over,} \\
& \emph{helped}      & \emph{and reading} & \emph{and skipping}  & \emph{and favoring it}   & \emph{and passing it over} \\
& \emph{the target}  & \emph{me helps}    & \emph{me helps}      & \emph{helps}             & \emph{helps} \\
\bottomrule
\end{tabular}
\end{table}

\begin{table}[htbp]
\centering
\small
\setlength{\tabcolsep}{3pt}
\caption{Readings of a \textbf{negative} attribution: the mirror of \cref{tab:signs}, where
demand and supply disagree. Note that within one neighborhood
$\sum_{j\in\N(n)} r^{\mathrm{att}}_{nj}=0$, so a node acting as destination always faces
demands of both signs.}
\label{tab:signtable}
\begin{tabular}{@{}lccccc@{}}
\toprule
& value path & \multicolumn{4}{c}{attention path} \\
\cmidrule(lr){2-2}\cmidrule(l){3-6}
& & \multicolumn{2}{c}{$n$ as source} & \multicolumn{2}{c}{$n$ as destination} \\
\cmidrule(lr){3-4}\cmidrule(l){5-6}
edge term & $\alpha_{in}\,g_i^\top m_n$
          & \multicolumn{2}{c}{$r^{\mathrm{att}}_{in}\,u_{in}$}
          & \multicolumn{2}{c}{$r^{\mathrm{att}}_{nj}\,v_{nj}$} \\
\midrule
supply (below) & ---
  & $u_{in}>0$ & $u_{in}<0$
  & $v_{nj}>0$ & $v_{nj}<0$ \\
 & & $n$ pushes logit up & $n$ pushes logit down
   & $n$ rates $j$ above & $n$ rates $j$ below \\
demand (above) & ---
  & $r^{\mathrm{att}}_{in}<0$ & $r^{\mathrm{att}}_{in}>0$
  & $r^{\mathrm{att}}_{nj}<0$ & $r^{\mathrm{att}}_{nj}>0$ \\
 & & score edge down & score edge up & shift away from $j$ & shift toward $j$ \\
\midrule
reading
& \emph{my content} & \emph{read me,}    & \emph{skip me,}     & \emph{I favor this one,} & \emph{I pass this one over,} \\
& \emph{hurt}       & \emph{but reading} & \emph{but skipping} & \emph{but favoring it}   & \emph{but passing it over} \\
& \emph{the target} & \emph{me hurts}    & \emph{me hurts}     & \emph{hurts}             & \emph{hurts} \\
\bottomrule
\end{tabular}
\end{table}

In the combined AttnLRP map, the attribution of a single node sums three components: the value path, and the source and destination roles of the attention path (\cref{tab:signs,tab:signtable}).
A single value entangles them, and cancellation can make a node whose components conflict indistinguishable from one with little influence.
Each role is itself a sum over edges and can cancel internally, as in any attribution that sums over features. Within one role, however, the summed relevance keeps a single, clear reading for the node, whereas cancellation across paths and roles mixes different interpretable readings. 

In general, the granularity of an attribution should follow the question asked of it.
The case study in \cref{sec:experiments} shows the \emph{skip me, and skipping me helps} reading of a positive attention path attribution, because the source role dominates. The attribution is positive because the demand on the anchor's outgoing logits is negative, and its features push these logits down. This lowers the attention its neighbors pay to it and limits the impact of its negative content. If neither role dominates, the roles may have to be interpreted separately. Otherwise, the attention attribution can again hide canceling effects, although they can both be attributed to routing influences.

All sign combinations across the value path and the two attention roles can occur, because no component constrains the sign of another. First, the value path is measured against a zero baseline, whereas the attention path demand is measured against the neighborhood, whose extent depends on the role. In the destination role, $n$'s features steer which of its 1-hop neighbors' messages it reads, and $n$ receives positive relevance if it rates neighbors with above-average messages higher and those with below-average messages lower. In the source role, the 2-hop neighborhood matters, because each destination evaluates $n$ relative to its own neighborhood. Thus, content that hurts the target can still receive positive demand if its destinations' other neighbors carry even less valuable content, and helpful content can receive negative demand if the competing content is more valuable. Second, value and supply project the same message $W_l x_n$ onto different directions ($g_i$ vs.\ $p_{in}$), so their signs are not constrained either. \cref{fig:bahex} shows both agreement (hubs) and disagreement (node $16$).

\section{Experimental setup}
\label{app:setup}
All models were implemented in PyTorch \citep{paszke2019pytorch} using PyTorch Geometric \citep{fey2019PyG,fey2025PyG} for message passing. NetworkX \citep{hagberg2008networkX} was used for graph generation and RDKit \citep{rdkit} for molecular featurization. All experiments were run on CPU (Apple M4). Training a single model takes approximately a couple of seconds for \textsc{BA-House} and \textsc{QKV-BA} and up to $1.5$ minutes on \textsc{ESOL}.

Each synthetic \textsc{BA} dataset is generated once and shared by all five runs per setting. These five seeds are for the model initialization only and not for data generation. Each attention-free control is trained under the same protocol as the model it controls for.

Layer hyperparameters are PyTorch Geometric defaults, with four exceptions. We use a single attention head throughout. Self-loops are written into the edge set rather than added inside the layer (\texttt{add\_self\_loops=False}), so that a loop is an ordinary edge with an ordinary attention weight and the message-masking metrics of \cref{app:expfid} operate on exactly the edge set the model sees. Source and destination projections are kept separate (\texttt{share\_weights=False}).
For the \texttt{TransformerConv} control we disable the root weight (\texttt{root\_weight=False}) and remove the unused \texttt{lin\_skip}. The LeakyReLU slope is $0.2$. No attention dropout is used; feature dropout is $0.1$ between the \textsc{ESOL} convolutions and $0$ elsewhere.

Datasets, architectures, training schedules and the set of attributed graphs are summarized in \cref{tab:datasets}.

\begin{table}[h]
\centering\small\setlength{\tabcolsep}{3pt}
\caption{Datasets, models and evaluation. All models share one skeleton: $L$ message-passing
layers of width $h$, one head, ReLU between convolutions, linear readout, sum-pooled where the
task is graph-level. \textsc{QKV-BA} has no activation after the last convolution, its target
being linear in $v$.}
\label{tab:datasets}
\begin{tabular}{@{}p{1.75cm}p{3.55cm}p{3.55cm}p{3.55cm}@{}}
\toprule
 & \textsc{QKV-BA} & \textsc{BA-House} & \textsc{ESOL} \\
\midrule
graphs & BA $n{=}8$, $m{=}2$
       & pos.: BA $n{=}15$, $m{=}2$ ${+}$ \textsc{HouseMotif}; neg.: BA $n{=}20$, $m{=}2$; balanced
       & $1116$ Delaney molecules ($1128$ before de-duplication) \\
node feat. & $x_i=[q_i|k_i|v_i]\in\R^{15}$, $d_q{=}d_k{=}d_v{=}5$, iid $U(-1,1)$
           & one-hot true degree, $10$ bins (loops excluded from the degree)
           & $35$-dim one-hot (element, degree, charge, $n$H, hybrid., aromaticity, ring) \\
edge feat. & --- & --- & $9$-dim one-hot bond \\
task & regress one Transformer head (below), MSE & graph classification, cross-entropy
     & graph regression on standardized $\log S$, MSE \\
model & GATv2 $h{=}16$, $L{=}1$, $629$ par. & GATv2 $h{=}32$, $L{=}1$, $834$ par.
      & GATv2 $h{=}64$, $L{=}4$, $32449$ par. \\
training & Adam, lr $10^{-2}$, cosine, $60$ epochs, batch $64$
         & Adam, lr $10^{-2}$, cosine, $30$ epochs, batch $32$
         & AdamW, lr $10^{-3}$, cosine, $600$ epochs, batch $32$, best-validation weights restored \\
split & $2000$ train / $500$ held-out, independently generated (seeds $0$ and $10000$)
      & $600$ train / $1200$ held-out, independently generated (seeds $0$ and $10000$)
      & $892/111/113$, one fixed random split (seed $0$) \\
attributed on & $60$ further graphs (seed $99$), node $0$ only
              & the $600$ positive held-out graphs, $\times5$ seeds $=3000$ records; per seed $248$ degree-four and $352$ degree-three anchors
              & all $1116$ molecules \\
explained scalar & $\sum$ of node $0$'s output & sum-pooled logit of the predicted class
                 & standardized output \\
\bottomrule
\end{tabular}
\end{table}

Nothing is selected on the \textsc{QKV-BA} and \textsc{BA-House} held-out sets: the epoch budget is fixed, with no early stopping and no best-weight restore, and the sets are independently generated rather than partitioned, so they serve as test sets. \textsc{ESOL} is the only model with selection, and reports on a test split untouched by it; its split is random rather than scaffold-based.
On \textsc{BA-House}, the explained class is the model's \texttt{argmax}, and no attributed graph is misclassified ($0$ of $3000$), so the predicted- and true-class conventions coincide.

\paragraph{\textsc{QKV-BA}} A constructed task where the ground truth of the split is known. Node features come in three blocks, $x_i=[q_i\,|\,k_i\,|\,v_i]$, and the target is one transformer attention head on the graph's neighborhoods,
\begin{equation*}
  y_i=\sum_{j\in\N(i)}\alpha_{ij}v_j,
  \qquad
  \alpha_i=\operatorname*{softmax}_{j\in\N(i)}\!\big(\beta\langle q_i,k_j\rangle/\sqrt{d_q}\big),
  \qquad \beta=8 .
\end{equation*}
Routing depends on $q$ and $k$ alone, content on $v$ alone. A model that fits this target must read $q,k$ in its scorer and $v$ in its messages, so a correct path split has to put value path relevance on $v$ and attention path relevance on $q,k$: the answer is fixed by the construction, not by our reading of it. Three constants make the task genuinely require attention rather than merely permit it.
\begin{itemize}[leftmargin=*,itemsep=0pt,topsep=2pt]
  \item \emph{$\beta=8$} sets how far the softmax commits. At low $\beta$ the weights are near-uniform and the target collapses towards a plain mean of the neighbors' $v$, which any fixed-weight aggregator can fit --- at $\beta=1$ a mean aggregator already reaches $R^2=0.93$, leaving almost nothing for routing to explain. At $\beta=8$ the head commits (normalized attention entropy $0.46$) and the best attention-free predictor reaches only $R^2=0.415$, so most of the target is reachable exclusively through input-dependent routing.
  \item \emph{$m\ge2$} makes neighborhoods overlap: two nodes share two non-self neighbors in $99.8\%$ of graphs. A static scorer (GATv1) ranks all sources in the same order regardless of who reads them \citep{brody2022gat2}, and so cannot rank the same two sources differently in two overlapping neighborhoods. The task therefore needs the dynamic scorer of GATv2.
  \item \emph{Separate $q$ and $k$ blocks} keep a node's score for itself an ordinary random draw, so the self-loop does not turn ``copy $v_i$'' into a shortcut.
\end{itemize}
Finally $h>d_v$, so that the readout is not a square, generically invertible map, which would force the value path onto $v$ by construction rather than by learning; the sweep $h\in\{5,10,16\}$ is reported in \cref{tab:relevance_mass_blocks}. One limit is inherent to the construction: $\alpha_i$ has $\deg_i-1$ free parameters against the $d_v$ equations $y_i$ supplies, so routing is identifiable only where $\deg_i-1\le d_v$.

\paragraph{\textsc{BA-House}} PyG's \textsc{HouseMotif} (eaves $0\!-\!1$, floors $2\!-\!3$, square $0\!-\!1\!-\!2\!-\!3$, rooftop $4$) is wired in by one edge from a uniformly random house node to a uniformly random main-graph node. Its house-side endpoint is the \emph{anchor}: attaching at an eave moves its degree $3\!\to\!4$ (\emph{degree-four anchor}), at a floor or the rooftop $2\!\to\!3$ (\emph{degree-three anchor}). The (before, after) pair is stored per graph, so statistics can be stratified without re-deriving the geometry.

\paragraph{\textsc{ESOL}} A real molecular regression: aqueous solubility ($\log S$) of
$1116$ small molecules, one graph per molecule, atoms as nodes and bonds as edges. Atoms carry a $35$-dimensional one-hot of chemical facts (element, degree, formal charge, attached
hydrogens, hybridization, aromaticity, ring membership), bonds a $9$-dimensional one (bond
type, conjugation, self-loop). There is no input encoder, so every input coordinate stays a
named chemical fact and relevance can be read per chemical property rather than per latent
dimension; eight of the $35$ coordinates are all-zero on this dataset and necessarily receive zero relevance. It is also the only dataset here with edge features, which in GATv2 enter the score and never the message (\cref{eq:gatv2}).

\subsection{Computing the two paths}
\label{app:impl}

\cref{eq:split} scales the gradient entering each factor of the product $\alpha_{ij}m_j$:
$w_{\mathrm{att}}$ into $\alpha$, $w_{\mathrm{val}}$ into $m$. Two implementations are
available. A custom backward per operation, as in AttnLRP \citep{achtibat2024AttnLRP}, evaluates
the product once but has to be hand-written. A reweighted forward, as in the detach form of
CP-LRP \citep{ali2022CPLRP}, rewrites the aggregation as
\begin{equation}
  \begin{split}
    y_i \;=\; &w_{\mathrm{att}}\sum_{j\in\N(i)}\alpha_{ij}\,[m_j]\texttt{.detach()}
    \;+\;w_{\mathrm{val}}\sum_{j\in\N(i)}[\alpha_{ij}]\texttt{.detach()}\;m_j\\[2pt]
    &+\;(1-w_{\mathrm{att}}-w_{\mathrm{val}})
      \Big[\textstyle\sum_{j\in\N(i)}\alpha_{ij}m_j\Big]\texttt{.detach()},
  \end{split}
  \label{eq:reweight}
\end{equation}
which is their trick with the weights freed --- CP-LRP is $\mathbf{w}=(1,0)$, where only the
middle term survives. The three terms sum to $y_i$, so the prediction never moves and only the backward is reweighted. We use this second form: stock autograd, one code path for every
$\mathbf{w}$, at the price of three product evaluations. The two agree to float precision.

Gradient\,$\times$\,Input on the reweighted forward gives G$\times$I at $\mathbf{w}=(1,1)$,
CP-LRP at $(1,0)$, AttnLRP at $(\tfrac12,\tfrac12)$ (identity rule at the activation) and the attention path at $(0,1)$. The split is taken at layer $1$ throughout.

\section{Further experiments}
\label{app:blind}
\begin{table}[H]
    \centering
    \caption{QKV-BA: Mean fraction of $|R|$ mass per block in node 0.}
    \label{tab:relevance_mass_blocks}
    \begin{tabular}{lccc}
        \toprule
        Model & $R^2$ & Value path (q/k/v) & Attention path (q/k/v) \\
        \midrule
        TransformerConv h=5   & 1.000 & 0.000 / 0.000 / 1.000 & 0.323 / 0.677 / 0.000 \\
        GATv2 h=5   & 0.413 & 0.019 / 0.043 / 0.939 & 0.168 / 0.166 / 0.666 \\
        GATv2 h=10  & 0.780 & 0.008 / 0.019 / 0.974 & 0.303 / 0.611 / 0.087 \\
        GATv2 h=16  & 0.950 & 0.005 / 0.003 / 0.992 & 0.326 / 0.610 / 0.064 \\
        \bottomrule
    \end{tabular}
\end{table}

\begin{figure}[h]
  \centering
  \includegraphics[width=0.8\textwidth]{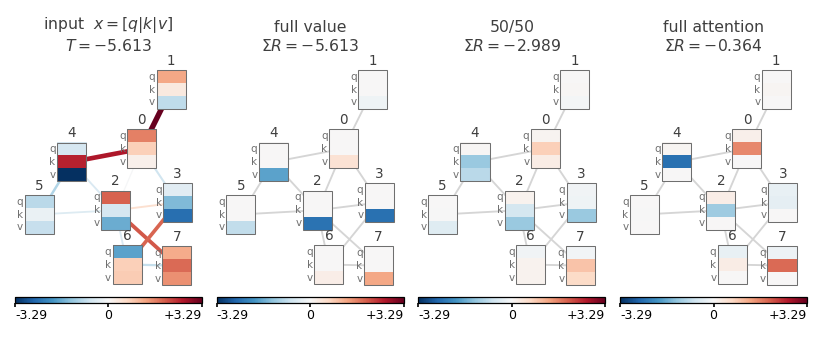}
  \caption{Shows separation of attributions by input block on separate path explanations.}
  \label{fig:qkv-ex}
\end{figure}

\paragraph{\textsc{QKV-BA}: each path is blind to what only the other one reaches} On \textsc{QKV-BA}, the input $x_i=[q_i|k_i|v_i]$ is blocked such that the target uses $q,k$ only for routing and $v$ only for content. The path split recovers this partition (\cref{tab:relevance_mass_blocks}, \cref{fig:qkv-ex}): a well-fitting GATv2 places $0.992$ of its value path mass on the value block and $0.936$ of its attention path mass on query and key. The separation is not exact because the forward function is not learned perfectly, some features from routing blocks reach the value block, and vice versa. Under-trained models thus separate less cleanly. A GATv2 with Transformer attention (\texttt{TransformerConv}), whose forward pass \emph{is} the target after the linear projections are learned, separates perfectly. 

\paragraph{\textsc{ESOL}: a molecular example} Edge attributes enter only the attention path by definition (\cref{eq:gatv2}) and therefore obtain no relevance in pure value path attributions. This is a structural property, which we confirm empirically on ESOL: the output has a nonzero gradient with respect to the edge attributes, whereas CP-LRP assigns them an exact zero. 

\begin{figure}[h]
  \centering
  \includegraphics[width=0.8\textwidth]{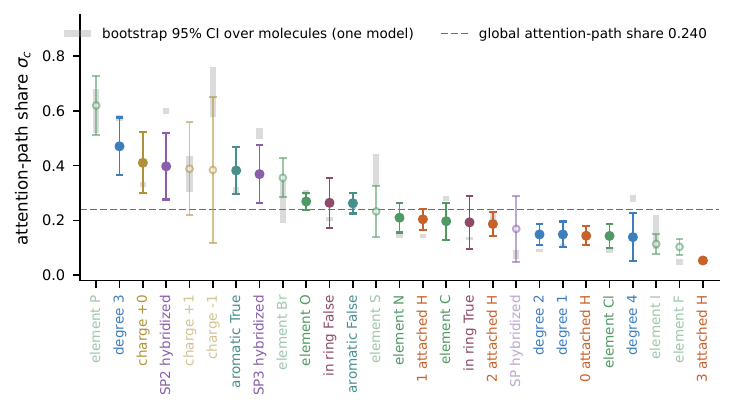}
  \caption{\textsc{ESOL}: attention-path share $\sigma_c=|\Ratt_{,c}|/(|\Ratt_{,c}|+|\Rval_{,c}|)$ per live input coordinate ($27$ of $35$; the other $8$ are always zero on \textsc{ESOL}), at the layer-1 split with the value path in all layers above, attributed over all $1116$ molecules and sorted by $\sigma_c$. Markers: mean $\pm$ SD over 5 initializations. Grey bars: $95\%$ bootstrap interval over molecules ($2000$ resamples) for the best-validation model, centered on that model's own estimate. Dashed line: global attention-path share, averaged over initializations ($0.240$; per initialization $0.201$--$0.280$). Hollow markers: coordinates present on fewer than $300$ atoms, excluded from all statistics.}
  \label{fig:esol-sigma}
\end{figure}

\cref{fig:esol-sigma} illustrates, on real-world data, that some input features receive proportionally more attention path relevance than the average and others less.
The pattern does not depend on which molecules are evaluated: computing $\sigma_c$ separately on two disjoint halves of the molecules gives almost the same ordering (Spearman $\rho=0.98$). It also largely persists across trainings, with an average Spearman $\rho=0.72$ over the $10$ pairs of initializations (range $0.61$--$0.87$). In every initialization, degree~3, charge~$+0$, SP2 and SP3 hybridization and aromaticity lean towards the attention path, while $0$, $2$ or $3$ attached hydrogens, degrees~1 and~2 and chlorine lean towards the value path. The figure motivates real-world applications in which mechanistic understanding could contribute to scientific discovery.

\begin{figure}[h]
  \centering
  \includegraphics[width=0.8\textwidth]{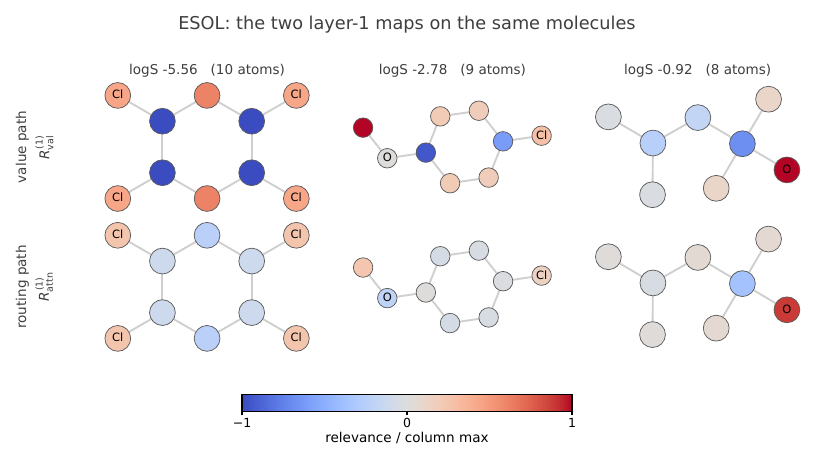}
  \caption{Three molecules spanning the solubility range. Top row: the value map summed over
  features per atom. Bottom row: the routing map, same atoms, same layer-1 split. Color is signed relevance normalized by the \emph{column} maximum --- value against routing on the same molecule is the comparison that is meaningful; magnitudes are not comparable across columns. Molecules are chosen deterministically for legibility, not for what they show.}
  \label{fig:esol-example}
\end{figure}
\cref{fig:esol-example} is a further, anecdotal illustration that the two paths pick up distinct information.
However, node-level agreement between the two paths does not reproduce across initializations in our experiments ($\rho\approx0.02$), only at the feature level. Future work could investigate input-feature ablations that induce stronger weighting of edge embeddings, or a learned embedding MLP on the chemical data, so that the network can form features suited to the routing and content roles of the message passing layer, which could be a step towards node-level agreement. Propagating relevance through such an embedding MLP would still recover the chemical feature relevances.

\section{Additional case study visualization}
\label{app:fullbahdegree}
\begin{figure}[H]
    \centering
    \includegraphics[width=0.8\textwidth]{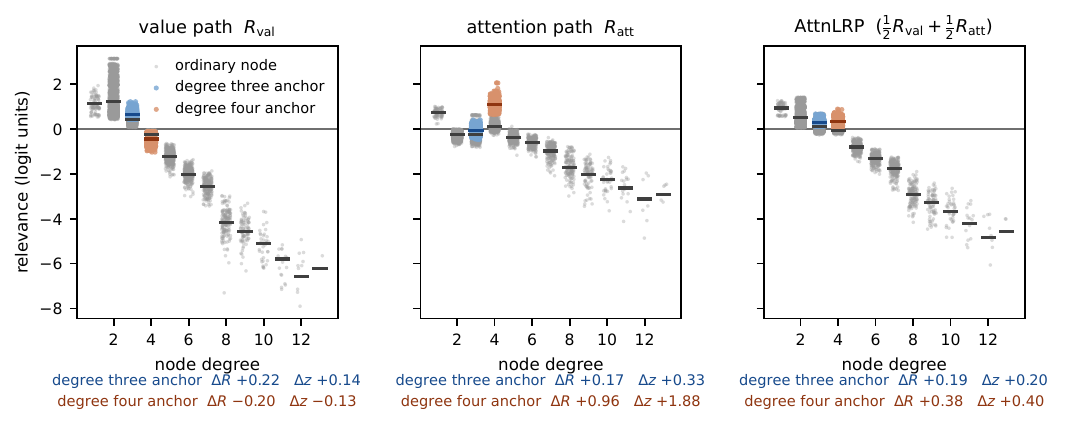}
    \caption{Node relevance for the house logit against degree, full version of \cref{fig:bahdegree}.}
    \label{fig:bahdegreefull}
\end{figure}

\section{Architecture control}
\label{app:bahctl}

The two attention architectures build their score differently: GATv2 adds a source and a destination projection, passes the sum through a LeakyReLU, and reuses the source projection as the message; \texttt{TransformerConv} takes a dot product of separate query and key projections. What they share is a bilinear product of attention weight and message under a softmax.

\begin{figure}[htbp]
  \centering
  \includegraphics[width=0.8\textwidth]{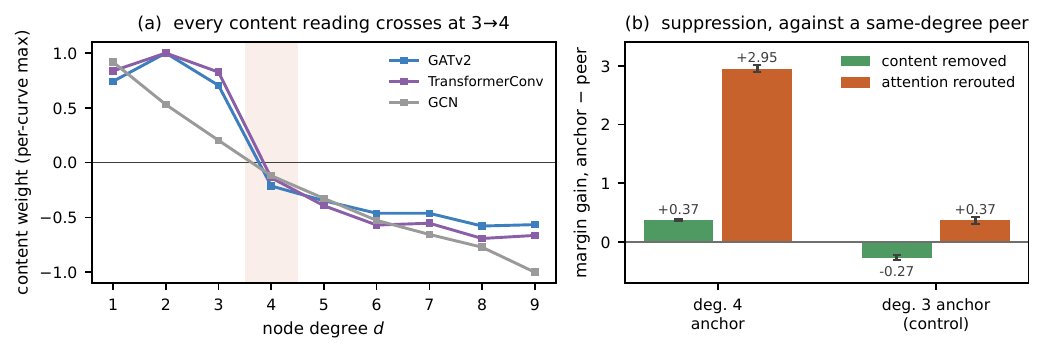}
  \caption{\textbf{(a)} Every content reading crosses between degree three and four; each curve is divided by its own largest value, since only the ordering and the sign boundary are the claim. \textbf{(b)} Margin gain from suppressing the anchor \emph{minus} the same quantity for an ordinary node of the same degree in the same graph, split into what removing content buys and what rerouting the freed attention buys on top. Degree-three anchors are the control. Bars are $95\%$ cluster-bootstrap intervals.}
  \label{fig:bahctl}
\end{figure}

Trained under the identical protocol, they behave alike. All three architectures, the attention-free GCN included, learn the same content reading: its sign changes between degree three and degree four in every architecture and seed combination (\cref{fig:bahctl}a). The routing reading is the part that is attention-specific --- the degree-four anchor separation is present for GATv2 and \texttt{TransformerConv}, and absent for the GCN. The shared source projection is therefore not what produces the reading, and the remaining differences in embedding and score do not change it, at least on this dataset.

\section{What metrics reward}
\label{app:expfid}

\begin{figure}[h]
    \centering
    \includegraphics[width=0.8\textwidth]{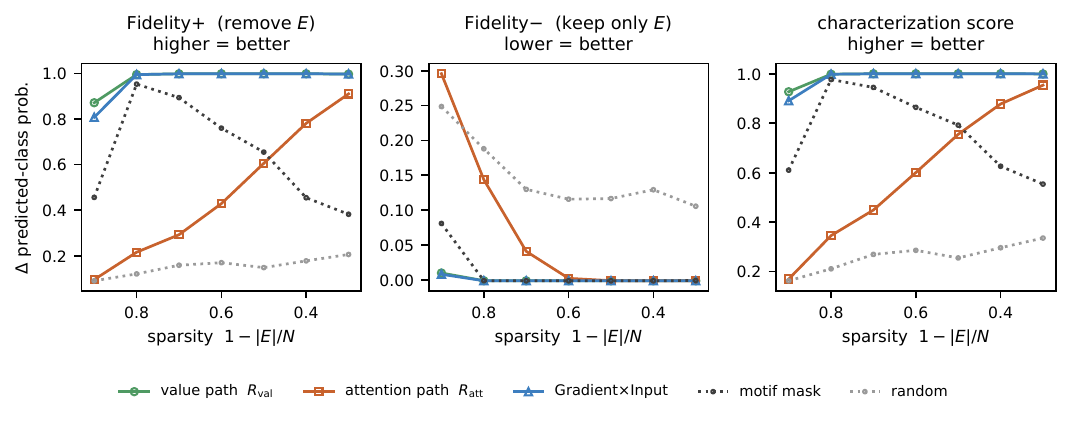}
    \caption{Fidelity metrics of \citet{yuan2023gnnxaisurvey}, with Fid+ and sparsity going back to \citet{pope2019xgcn}, which remove a node by masking the messages from and to it. Fid+: drop in predicted-class probability when the explanation is removed, so high means necessary. Fid-: the same drop when \emph{only} the explanation is kept, so low means sufficient. Sparsity is the fraction of the graph not in the explanation; characterization score \citep{amara2022GraphFramEx} is the harmonic mean of Fid+ and (1 - Fid-).}
    \label{fig:bahfid}
\end{figure}

\cref{fig:bahfid} places the value path near the ceiling ($\mathrm{Fid+}=0.98$, char.\ $0.99$) and the attention path far behind it ($0.50$, char.\ $0.61$), still outperforming random.
Fidelity+ cannot measure the suppression-based positive evidence, it only rewards removals that lower the prediction. On node 16 of \cref{fig:bahex}, masking its edges without renormalizing raises the logit margin by $+1.23$, and letting the softmax redistribute the attention it freed raises it by $+4.39$. Either way, the model classifies the graph \emph{better} once it can no longer read from that node, and this holds for every $3\!\to\!4$ anchor node in the evaluation set. The gap between the value and attention paths reflects, at least in part, the question posed by the metric rather than simply the quality of the explanations.

\begin{figure}[h]
    \centering
    \includegraphics[width=0.4\textwidth]{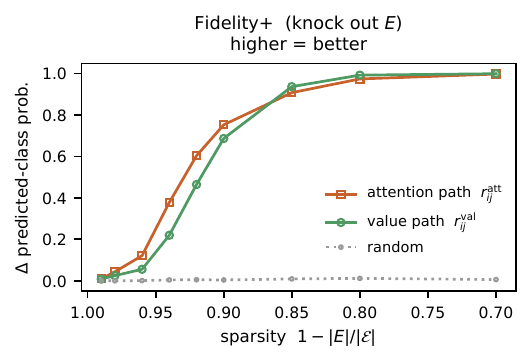}
    \caption{Fidelity+ with attention knockout instead of message masking: the most positive edges of each edge attribution map have their logits set to $-\infty$ together, allowing neighborhood attention to renormalize. Same models and graphs as \cref{fig:bahfid}. $\mathcal{E}$ is the set of all edges. Up to sparsity of 0.70: 30\% of edges removed.}
    \label{fig:bah_knock}
\end{figure}

In Transformer research, a useful tool for mechanistic interpretability next to value zeroing is \emph{attention knockout} \citep{geva2023dissecting}. It answers the question: what would happen if this logit did not exist? Deletion here does not mean zero but $s_{ij} \to - \infty$, because this lets the neighborhood distribute its attention as if the edge did not exist. Over all 12,900 edges of the graphs in \cref{fig:bahfid}, the logit relevance $r^{att}_{ij}$ (\cref{eq:ratt}) predicts single-edge knockout with Spearman $\rho = 0.99$, while the value path's edge relevance $r^{val}_{ij}$ reaches $\rho = 0.86$, because it measures an edge's content against zero rather than against the rest of its neighborhood. Used as a fidelity metric like \cref{fig:bahfid}, knocking out the most relevant edges of each map together (\cref{fig:bah_knock}), the attention path lowers the class probability faster while fewer edges are removed. Knocking out edges with highest content, those of the value path, catches up as the number of removed edges increases and the single-edge prediction no longer holds. 

That the attention path's logit relevance predicts the removal of edges well (\cref{fig:bah_knock}), while its node-level relevance predicts the removal of nodes poorly (\cref{fig:bahfid}), goes along with the demand supply interpretation (\cref{ssec:types}, \cref{app:flows}). Each node's features can contribute positively or negatively to a logit whose relevance already has a clear, individually signed role within its neighborhood.
These examples illustrate how the choice of evaluation metric affects the interpretation of explanation scores. Further perturbation research, especially at the node level, and an analysis of how to leverage disentangled attention path and value path attributions for downstream explanations are left for future work.

\end{document}